# Impact of Latent Space Dimension on IoT Botnet Detection Performance: VAE-Encoder Versus ViT-Encoder

Hassan Wasswa
*School of Systems and Computing*
*University of New South Wales*
Canberra, Australia
h.wasswa@unsw.edu.au

Aziida Nanyonga
*School of Engineering and Technology*
*University of New South Wales*
Canberra, Australia
a.nanyonga@unsw.edu.au

Timothy Lynar
*School of Systems and Computing*
*University of New South Wales*
Canberra, Australia
t.lynar@unsw.edu.au

*Abstract*—The rapid evolution of Internet of Things (IoT) technology has led to a significant increase in the number of IoT devices, applications, and services. This surge in IoT devices, along with their widespread presence, has made them a prime target for various cyber-attacks, particularly through IoT botnets. As a result, security has become a major concern within the IoT ecosystem. This study focuses on investigating how the latent dimension impacts the performance of different deep learning classifiers when trained on latent vector representations of the train dataset. The primary objective is to compare the outcomes of these models when encoder components from two cutting-edge architectures: the Vision Transformer (ViT) and the Variational Auto-Encoder (VAE) are utilized to project the high dimensional train dataset to the learned low dimensional latent space. The encoder components are employed to project high-dimensional structured .csv IoT botnet traffic datasets to various latent sizes. Evaluated on N-BaIoT and CICIoT2022 datasets, findings reveal that VAE-encoder based dimension reduction outperforms ViT-encoder based dimension reduction for both datasets in terms of four performance metrics including accuracy, precision, recall, and F1-score for all models which can be attributed to absence of spatial patterns in the datasets the ViT model attempts to learn and extract from image instances.

*Index Terms*—IoT botnet, Vision transformer, Self-attention, Variational Auto-encoder, Attack detection

## I. INTRODUCTION

Information security is of paramount importance and has become a core requirement for doing any kind of business across the globe. Currently, information systems are extensively interconnected and harbour all kinds of confidential information at both personal and organisational level. However, the past decade has seen an explosion of a multitude of cyber-attack avenues ranging from digital image spoofing-based attacks to use of IoT sensors [1]–[3]. Consequently, the ability to precisely and efficiently detect cyber-attacks in the current highly dynamic attack surface driven by the rapidly evolving IoT technology is of utmost importance for realizing the CIA (Confidentiality, Integrity, and Availability) triad of information assurance. However, despite its enormous benefits, IoT's security dimensions do not match the technology's growth rate making it a potent tool for various kinds of cyber-attacks notably IoT botnet-based attacks [4], [5]. Various studies have demonstrated different attacks that can be realized through use of IoT devices/sensors [2], [6]. This can be attributed to IoT technology deploying devices that are highly limited in resources including computing power, memory, and power supply, consequently making them unfit for running legacy defense schemes like conventional antivirus software, intrusion detection systems, etc [4], [5].

To ameliorate this challenge, extensive research has been conducted aiming to propose efficient models that can support IoT-based attack detection. A common approach for enhancing the detection performance of learning models is by training them within a reduced dimension space. A number of prior studies have demonstrated that working in a low dimensional space comes with benefits such as reduced system requirements, and enhanced model performance [7]–[9]. This is because the dimension reduction process aims to extract only fields that are sufficient for traffic discrimination. The most recent dimension reduction technique that has received extensive attention is the Auto-encoder (AE) [4], [10] and its variants such as Variational Auto-encoder (VAE) [11]. Using the encoder component of a pre-trained AE/VAE, the high dimensional train set is projected to a low dimensional latent space. Classifiers are then trained on the learned latent space representations. This research's aim is to investigate the impact of latent space size on the classification performance of different deep learning models when two state-of-the-art encoder-based schemes, (including the include VAE-based encoder and ViT-based encoder) are employed for dimension reduction.

Despite the two architectures having recorded out-standing performance in various AI-based studies across a multitude of research problems [12]–[17], no recent study has given a comprehensive analysis in terms of how deep learning models compare when trained on latent space representations from the two models. This study aimed at closing this knowledge gap by varying the latent space dimension using both VAE and ViT while training five deep learning classifiers on two well investigated IoT botnet datasets (N-BaIoT and CICIoT2022 datasets). Model performance was measured using four princi-ple performance metrics including accuracy (ACC), Precision (Prc), Recall (Rec) and F1-score (F1). The contribution of this work is three-fold:

1) We employed the ViT-based model, a model that was purposely designed for image and video data classification, for classification of structured IoT traffic in form of .csv files.
2) We explore the impact of various latent space dimensions from both a classical VAE-encoder and ViT encoder on the classification performance of deep learning models.
3) The study compares the performance of ViT-based models against VAE-based models in classification of IoT botnet traffic.

The remainder of this work is structured as follows: section II, gives an overview of the existing research on IoT botnet detection. Section III provides a comprehensive description of the approach deployed in this study, along with details of used datasets and the various methods employed. Section IV is dedicated to presenting and discussing the study's results, with the conclusion presented in Section V.

## II. RELATED WORK

We discuss this work in the context of the following two bodies of the state-of-the-art: (1) prior studies deploying VAE for attack detection, and, (2) prior studies deploying Transformers for attack detection.

### A. Prior prior studies deploying Transformers for attack detection

Because the transformer model was originally designed for NLP tasks, where training features show sequential patterns, and the ViT was initially created for image and video data, which exhibit spatial structures, transformer models have not seen extensive application in the field of attack detection with regard to captured net flow packets. This limitation primarily arises from the lack of significant sequential patterns or spatial structures by features extracted from network packet captures conventionally do not display.

The study in [18] introduced two novel self-attention-transformer-based models, SeqConvAttn and ImgConvAttn, as alternatives to the CNN-based classifiers for malware classification. The authors then merged the two models to proposed a two-stage framework that took file size into account with focus on balancing the trade-off between accuracy and classification latency. By training the models on three malware datasets, the authors demonstrates that transformer-based models can attain higher classification accuracy compared to conventional CNN-based designs.

The ViT model was employed in study [19] to extract fine facial attributes from public face image spoofing datasets, including spoofing in the wild (SiW) [20] and Oulu-NPU [21] datasets, for image spoofing attack detection. To enhance the detection of unseen images, three forms of patch-wise data augmentation were suggested and implemented, including intraclass patch mixing, live patch masking, and patch cutout data augmentation. This approach outperformed existing state-of-the-art methods in terms of APCER, BPCER, and ACER.

In study [22], an approach was introduced for classifying IoT botnet malware using a transformer model. The experimental study involved disassembling the malware opcodes of two well-known malware classes, Mirai and Gafgyt (commonly known as BASHLITE), one unknown IoT malware, and benign samples. By implementing a script, instruction sequences of the opcode functions were extracted, and CBOW was used to create word vector representations. The proposed transformer model was then employed to extract features, and a feed-forward neural network was trained for classification.

In study [23], a scheme for detecting domain generation algorithm (DGA) domain names, often used by botnet devices to communicate with the command-and-control (C&C) server in a botnet attack, was proposed. The pretrained CANINE transformer-based architecture was used to distinguish between benign domain names and DGA-generated domain names. To enable the CANINE transformer model to process the domains, each domain name was tokenized into individual character sequences, and the three required input components (input IDs, attention mask, and token type ID) were created and employed to fine-tune the transformer model.

In study [24], a method was proposed for detecting cross-site scripting (XSS) attacks, and the performance of a transformer architecture-based model was compared with two RNN models, including LSTM and GRU. Each URL was broken down into its constituent text components to create sequences of words. In order to train the transformer, various natural language processing techniques, such as tokenization, word2vec with a skip-gram model, TF-IDF, and CBOW, were utilized to generate and represent word feature vectors. The authors reported that the Transformer model had the potential to enhance the accuracy of XSS attack detection.

### B. Prior studies deploying Variational Auto-encoders for attack detection

As a generative model, the VAE has mostly been utilized for data augmentation within imbalanced learning settings to enhance classification performance with limited application to dimension reduction.

In study [5], a novel three-phase methodology was demonstrated, employing auto-encoders, VAEs, and cost-sensitive learning techniques. The primary objective was to improve the recognition of underrepresented attack classes in highly imbalanced IoT traffic datasets. The method involved generating additional instances of the minority class in a controlled manner while ensuring that the synthetic instances did not outnumber the primary observations of the minority classes. The approach exhibited exceptional performance when evaluated on the BoT-IoT and CICIoT2022 datasets.

In [12], a strategy for intrusion detection was proposed that leveraged a VAE to generate more instances of the attack class, aiming to create a balanced dataset for intrusion detection. The VAE was trained using HDFS dataset log files and TTP dataset data from a private industry. A multilayer perceptron was subsequently trained for attack detection. Additionally, the proposed model incorporated a range-based sequential search algorithm within the detection engine for data segmentation.

In study [13], a conditional VAE was trained on the NSL-KDD dataset to address its inherent class imbalance. This approach utilized the Laplace distribution to approximate the posterior distribution, with instance labels provided as inputs to enable conditional learning. A Deep Neural Network (DNN) was then trained on both the original unbalanced dataset and a balanced dataset containing synthetic samples, allowing for performance comparison. In study [25], a VAE was trained on the minority class of the NSL-KDD dataset, and synthetic samples were generated for the minority class to mitigate the class imbalance in the dataset's distribution.

## III. METHODOLOGY

This work deployed the ViT model for classification of IoT traffic on two datasets. The process involved extracting .csv instances from IoT traffic .pcap files and reshaping them as one-channel 2D images before feeding them into the ViT encoder model for feature learning. The ViT encoders' low dimensional output was then fed into a deep learning algorithm for classification. The performance of five deep learning models (Section III-E) was evaluated when trained on latent space vector representations of dimension 2, 6, 10 and 14, in terms of Accuracy (ACC), Precision (Prc), Recall (Rec) and F1-score (F1), with the aim of exploring the impact of latent dimension on model performance. The study also deployed VAE-based models in which the respective encoder components were used to project the high dimensional train set features to low dimensional latent space representations. The five classifiers were again trained and evaluated for performance comparison. In simple terms, for the purpose of fair comparison, classification models in both cases were trained on low dimensional latent space vector representations from the respective encoder models. Figure 1 shows the two models' experimental flows: (a) ViT-based approach and (b) VAE-based approach. Details of the datasets, techniques and models are presented in subsequent subsections.

### A. Datasets

All models were trained and their performance evaluated on two well investigated public IoT traffic datasets: the N-BaIoT [4] and the CICIoT2022 [26] datasets. The N-BaIoT dataset constitutes 115 independent features that were extracted from netflow packets of a seed IoT network. The target class for this study is the "subcategory" column which categorizes each dataset instance as belonging to one of nine traffic classes including "normal", "Scan", "junk", "Combo", "tcp flood", "udp flood", "SYN flood", "ACK flood", "udp flood" and "udpplain" for multiclass classification.

On the other hand, the CICIoT2022 dataset, created by the Canadian Institute of Cybersecurity (CIC), contains benign and attack traffic captures in .pcap format. Employing the revised CICFlowMeter[1] tool, 87 traffic flow features were extracted, resulting in a dataset comprising over 3.2 million netflow records. The dataset's five multi-class labels (Normal, HTTP flood, TCP flood, UDP flood, and Brute force) were derived from directory names.

[1]https://github.com/GintsEngelen/CICFlowMeter

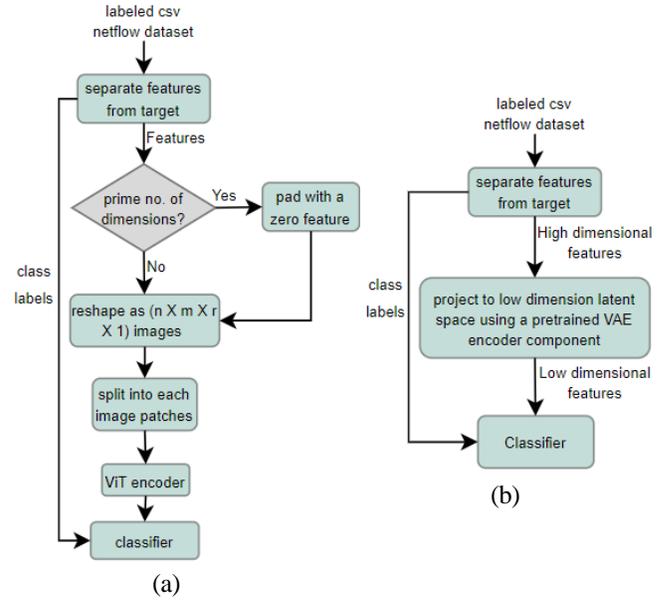

Fig. 1. Design flow of the two state-of-the-art model architectures: (a) Transformer-based (b) VAE-based models

### B. Data preprocessing

The data preprocessing involved removing missing values and dropping fields that were deemed irrelevant to model training. In both datasets, fields including IP addresses (both source and destination), packet number and sequence number/id were dropped. This was followed by reshaping each training instance to take the shape of a 1-channel 2D image. That is, each instance, $x \in \mathsf{R}^d$ is expressed as $x' \in \mathsf{R}^{r \times k \times 1}$ where $r \times k = d$. Each $x'$ is then split into equal sized image patches before it is fed into the ViT model. For cases where $d$ is a prime number, zero-padding was introduced increasing the dimension $d$ to $d + 1$ before the reshaping phase.

Since the N-BaIoT dataset constituted 115 independent variables, each instance was reshaped as a $5 \times 23 \times 1$ image and 23 equal-sized $5 \times 1 \times 1$ images patches were then created from each of the image-shaped instances. On the other hand, the CICIoT2022 dataset comprised 84 independent variables after cleaning and each instance was reshaped as a $6 \times 14 \times 1$ image. Each reshaped instance was then split into 21 equal-sized $2 \times 2 \times 1$ image patches.

### C. Transformer

The concept of Transformers was first presented in study [27], and brought about a revolutionary shift in the realm of NLP tasks by achieving exceptional outcomes. Contrast to conventional RNN models, the Transformer employs a multi-head self-attention mechanism to enable parallel processing consequently mitigating the slow sequential training limitations associated with conventional RNNs. Because of the benefits associated with the Transformer models, the study in [28] proposed the Vision Transformer (ViT) as as an extension of the self-attention mechanism introduced in the conventional transformer model, which was originally designed for NLP, to enable image processing in computer vision tasks. Unlike traditional CNNs, ViT works by dividing images into fixed-size patches followed by application of the self-attention mechanism to capture relationships among these patches. This enables ViT to efficiently understand global context and long-range dependencies in images. The fixed-sized patches are flattened into vectors and linearly embedded to create initial representations. To retain spatial information, ViT introduces positional embedding for each flattened patch vector enabling the model to understand the relative positions of patches within the image.

### D. Variational Autoencoder

The introduction of VAE [11] aimed to address the lack of regularization in the latent space, a common issue in traditional AE. VAE resolves this problem by imposing constraints on the distribution of the latent space, making it more akin to a normal distribution. It employs Bayesian variational inference to learn parameters for both the encoder and decoder. VAE approximates the true distribution of a dataset, $D$, by introducing a latent variable, $z$, and randomly sampling a variable, $x$, from $D$. This approach enables the learning of the joint distribution, $p_\psi(x, z)$, between $x$ and $z$ through estimating a set of parameters, $\psi$. Consequently, the marginal distribution, $p_\psi(x)$, can be expressed as Eq. 1.

$$p_\psi(x) = \int p_\psi(x,z)dz \quad (1)$$

Calculating $p_\psi(x)$ is challenging and computationally expensive in the VAE framework. To mitigate this, VAE approximates $p_\psi(x)$ by assuming a simpler distribution for the latent variable $z$, treating $p_\psi(x|z)$ as $p_\psi(z|x)$. This approximation facilitates the learning of parameters $\beta$ for a surrogate network $q_\beta(z|x)$, that estimates $p_\psi(z|x)$, and allows for the calculation of the expected value of $log p_\psi(x)$ over $\beta$ as shown in Eq. 2.

$$E_{z \sim q_\beta(z|x)} log p_\varphi(x) = L(\psi, \beta, z) + D_{KL}(q_\beta(z|x)||p_\psi(z|x)) \quad (2)$$

Where $L(\psi, \beta, z)$ represent the Evidence Lower Bound (ELBO), and $D_{KL}(q_\beta(z|x)||p_\psi(z|x))$ stand for the Kullback Leibler divergence, $(D_{KL})$, between the approximate and true posterior distributions. VAE aims to learn the parameters, $\beta$, which correspond to the decoder model's weights that approximate the posterior distribution, $q_\beta(z|x)$. Its training is focused on maximizing the ELBO and, consequently, minimizing $D_{KL}$, as the expectation remains constant for $x$. This facilitates the generation of new samples, denoted as $x'$, by drawing from the latent encoding, $z$, in such a way that $x$ is approximately equal to $x'$.

### E. Classification models

Five deep learning models including a standard feed-forward Deep Neural Network (DNN), LSTM, Bidirectional LSTM (BLSTM), Gated Recurrent Unit(GRU), and SimpleRNN (sRNN), were deployed for the classification phase. All models were implemented following an architecture of four hidden layers. The DNN model constituted four hidden dense layers while for RNN-based models including LSTM, BLSTM, GRU, and sRNN, the first hidden layer of DNN was substituted for by the respective model layer and the input reshaped accordingly. Rectified Linear unit (ReLu) was used for neuron activation except for the output layer where the softmax activation function was used. For all model training, the Adam optimizer was used while categorical-cross entropy was deployed as the loss function. We also added dropout layers to randomly drop 30% and 20% of the neuron outputs after the third and fourth hidden layers to reduce the chances of model over-fitting.

## IV. RESULTS AND DISCUSSION

The experimental findings from both datasets reveal that latent space dimension impacts the performance of the various models differently across different performance metrics.

### A. ViT-based classification performance from the N-BaIoT dataset

Table I presents the classification performance of the five deep learning models when the 115-dimensional N-BaIoT dataset is projected to latent spaces of dimension 2, 6, 10, and 14 using the ViT-encoder model. For all models, the performance, in terms of all metrics, is considerably low at a latent dimension of 2 and improves as the latent size grows up to size 10, at which point the models start to show no significant change in performance with latent size increase. This pattern shows that with only 2 latent features so much of information pertinent for model learning is lost hence the poor performance. On the other hand, at latent size 10 the models have sufficient information to precisely discriminate between the various traffic instances of the N-BaIoT dataset. The classification results also reveal that sRNN generally outperforms other models across all metrics at smaller latent space dimensions. The insignificant change in performance between latent dimension 10 and latent dimension 14 implies that, with the right parameter setting, a latent size of 10 is sufficient for the models to precisely discriminate between N-BaIoT traffic instances.

### B. VAE-based classification performance from the N-BaIoT dataset

The results in Table II reveal that, generally, the LSTM model outperforms the other four models with scores of 96.82, 0.95, 0.95, and 0.95 for accuracy, precision, recall, and F1 respectively, at latent dimension 10 while BLSTM performed best with scores of 97.51, 0.96, 0.93, and 0.93 at latent dimension 14. Like in the case of the ViT-based models, the VAE-based models recorded low performance at latent dimension 2 and the performance is seen to significantly rise until latent dimension 10. However, between latent dimension 10 and 14 the changes are generally insignificant.

TABLE I
CLASSIFIER PERFORMANCE ON N-BAIOT-VIT-ENCODED LATENT SPACE VECTORS

| Model | latent dim = 2 | | | | latent dim = 6 | | | | latent dim = 10 | | | | latent dim = 14 | | | |
|---|---|---|---|---|---|---|---|---|---|---|---|---|---|---|---|---|
| | Acc | Prc | Rec | F1 | Acc | Prc | Rec | F1 | Acc | Prc | Rec | F1 | Acc | Prc | Rec | F1 |
| DNN | 71.08 | 0.50 | 0.52 | 0.50 | 85.29 | 0.73 | 0.76 | 0.74 | 92.17 | 0.87 | 0.88 | 0.87 | **94.12** | **0.90** | 0.89 | **0.90** |
| LSTM | 73.65 | 0.63 | 0.55 | 0.52 | 91.86 | 0.77 | 0.81 | 0.79 | 92.54 | 0.87 | 0.88 | 0.87 | 89.82 | 0.85 | 0.87 | 0.85 |
| BLSTM | 69.55 | 0.46 | 0.55 | 0.48 | 87.67 | 0.84 | 0.83 | 0.83 | 89.20 | **0.89** | 0.77 | 0.78 | 93.82 | 0.89 | 0.89 | 0.89 |
| GRU | 75.49 | 0.51 | 0.53 | 0.51 | 88.07 | 0.85 | 0.83 | 0.83 | 90.59 | 0.88 | 0.85 | 0.85 | 93.77 | 0.89 | 0.89 | 0.89 |
| sRNN | **78.01** | **0.81** | **0.67** | **0.65** | **92.61** | **0.88** | **0.88** | **0.87** | **93.45** | **0.89** | **0.89** | **0.89** | 93.95 | **0.90** | **0.90** | 0.89 |

TABLE II
CLASSIFIER PERFORMANCE ON N-BAIOT-VAE-ENCODED LATENT SPACE VECTORS

| Model | latent dim = 2 | | | | latent dim = 6 | | | | latent dim = 10 | | | | latent dim = 14 | | | |
|---|---|---|---|---|---|---|---|---|---|---|---|---|---|---|---|---|
| | Acc | Prc | Rec | F1 | Acc | Prc | Rec | F1 | Acc | Prc | Rec | F1 | Acc | Prc | Rec | F1 |
| DNN | **83.69** | 0.76 | **0.72** | 0.67 | 94.56 | 0.86 | 0.83 | 0.81 | 95.58 | 0.93 | 0.93 | 0.93 | 94.10 | 0.80 | 0.83 | 0.81 |
| LSTM | 82.44 | 0.78 | 0.69 | 0.66 | 94.79 | 0.90 | 0.88 | 0.88 | **96.82** | **0.95** | **0.95** | **0.95** | 96.50 | 0.95 | 0.92 | 0.92 |
| BLSTM | 83.46 | **0.82** | **0.72** | **0.69** | 94.85 | 0.80 | 0.84 | 0.82 | 95.60 | **0.95** | 0.89 | 0.90 | **97.51** | **0.96** | **0.93** | **0.93** |
| GRU | 82.27 | 078 | 0.69 | 0.66 | **95.81** | **0.94** | **0.90** | **0.91** | 95.70 | 0.94 | 0.85 | 0.83 | 95.84 | 0.92 | 0.86 | 0.85 |
| sRNN | 83.40 | 0.68 | 0.71 | 0.67 | 94.97 | 0.86 | 0.84 | 0.82 | 94.47 | 0.94 | 0.84 | 0.84 | 96.22 | 0.93 | 0.88 | 0.87 |

TABLE III
CLASSIFIER PERFORMANCE ON CICIOT2022-VIT-ENCODED LATENT SPACE VECTORS

| Model | latent dim = 2 | | | | latent dim = 6 | | | | latent dim = 10 | | | | latent dim = 14 | | | |
|---|---|---|---|---|---|---|---|---|---|---|---|---|---|---|---|---|
| | Acc | Prc | Rec | F1 | Acc | Prc | Rec | F1 | Acc | Prc | Rec | F1 | Acc | Prc | Rec | F1 |
| DNN | 91.74 | 0.92 | 0.70 | 0.72 | 96.24 | 0.98 | 0.87 | 0.91 | 95.98 | 0.98 | 0.84 | 0.89 | 96.31 | 0.98 | 0.88 | 0.92 |
| LSTM | **94.50** | **0.97** | **0.75** | **0.81** | 96.28 | 0.98 | 0.87 | 0.91 | 96.26 | 0.98 | 0.88 | 0.91 | 96.41 | 0.98 | 0.88 | 0.92 |
| BLSTM | 93.48 | 0.93 | 0.71 | 0.74 | 96.22 | 0.98 | 0.87 | 0.91 | 96.20 | 0.97 | 0.88 | 0.91 | 96.33 | 0.98 | 0.88 | 0.92 |
| GRU | 93.63 | 0.77 | 0.67 | 0.70 | 96.29 | 0.98 | 0.87 | 0.91 | 96.28 | 0.98 | 0.87 | 0.91 | 96.34 | 0.98 | 0.88 | 0.91 |
| sRNN | 93.39 | 0.75 | 0.67 | 0.70 | 96.21 | 0.98 | 0.87 | 0.91 | 96.32 | 0.98 | 0.88 | 0.91 | 96.34 | 0.98 | 0.88 | 0.91 |

TABLE IV
CLASSIFIER PERFORMANCE ON CICIOT2022-VAE-ENCODED LATENT SPACE VECTORS

| Model | latent dim = 2 | | | | latent dim = 6 | | | | latent dim = 10 | | | | latent dim = 14 | | | |
|---|---|---|---|---|---|---|---|---|---|---|---|---|---|---|---|---|
| | Acc | Prc | Rec | F1 | Acc | Prc | Rec | F1 | Acc | Prc | Rec | F1 | Acc | Prc | Rec | F1 |
| DNN | **98.68** | 0.78 | **0.57** | **0.63** | 98.90 | 0.97 | 0.71 | 0.78 | 99.21 | 0.99 | 0.88 | 0.92 | 99.22 | 0.99 | 0.87 | 0.92 |
| LSTM | 98.61 | 0.59 | 0.50 | 0.52 | 99.20 | 0.98 | 0.88 | 0.91 | 99.23 | 0.99 | 0.88 | 0.92 | 99.24 | 0.99 | 0.88 | 0.92 |
| BLSTM | 98.60 | 0.58 | 0.50 | 0.53 | 99.21 | 0.98 | 0.88 | 0.91 | 99.22 | 0.99 | 0.88 | 0.92 | 99.24 | 0.99 | 0.88 | 0.92 |
| GRU | 98.59 | 0.58 | 0.50 | 0.53 | 99.21 | 0.98 | 0.88 | 0.92 | 99.22 | 0.98 | 0.88 | 0.92 | 99.25 | 0.99 | 0.88 | 0.92 |
| sRNN | 98.61 | **0.79** | 0.51 | 0.55 | 99.18 | 0.98 | 0.86 | 0.91 | 99.22 | 0.99 | 0.88 | 0.92 | 99.24 | 0.99 | 0.88 | 0.92 |

### C. ViT-based classification performance from the CICIoT2022 dataset

From the findings in Table III, LSTM outperformed all other models in terms of all metrics when the ViT encoder's latent dimension was set to 2. However, from latent dimension 6 onwards, no model shows superior performance over the other and no significant change of performance is recorded by any of the models from this point.

### D. VAE-based classification performance from the CICIoT2022 dataset

Unlike the case of ViT-based models where LSTM showed superior performance at latent dimension 2, for VAE-based models DNN outperformed all other models at this setting as can be seen in Table IV. However, model performance almost stabilized at latent dimension 6 with no particular model showing superior performance over other models and with no significant change.

### E. Comparative analysis of ViT versus VAE-based model performance

In general the VAE-based models outperformed the ViT-based models on both datasets across all performance metrics for all latent dimension sizes. This relatively low performance of ViT-based models can be attributed to the datasets' lack of spatial patterns for which the ViT model was initially designed to learn and extract from image and video datasets. However, because of the multi-head attention mechanism that allows parallel processing, the ViT-based models converge faster than the VAE-based models. For all models the ViT-based models reached the optimal performance within 5 epochs during training. Also, both models recorded better results when evaluated on CICIoT2022 dataset compared to N-BaIoT dataset. A number of factors including number of classes in the target class (N-BaIoT's 9 classes vs CICIoT2022's 5 classes), initial dataset dimension size (115-dimensional N-BaIoT vs 84-dimensional CICIoT2022), initial attribute-based statistical properties among others can account for this performance discrepancy. Overall, both models recorded commendable performance and ViT-based results showed that Transformers can potentially enhance IoT botnet attack detection with outstanding speeds when the right set of parameters is used.

## V. CONCLUSION

The widespread availability of IoT devices coupled with their inherent limitations has rendered them vulnerable to various cyber-attacks, especially through IoT botnets. Consequently, security has emerged as a prominent concern within the IoT ecosystem. This research was centred on exploring how latent dimension influences the performance of different deep learning classifiers when they are trained on latent vector representations of training datasets. The study compared the models' results when the ViT-encoder is utilized to transform high-dimensional structured .csv instances extracted from IoT net flow captures (.pcap) into various latent sizes against results when the VAE-encoder is used. When evaluated using the N-BaIoT and CICIoT2022 datasets, the findings indicate that dimension reduction with the VAE encoder outperforms the ViT encoder-based dimension reduction for both datasets across four performance metrics, including accuracy, precision, recall, and F1-score.